# Understanding and Exploiting Dependent Variables with Deep Metric Learning

Niall O' Mahony, Sean Campbell, Anderson Carvalho, Lenka Krpalkova, Gustavo Velasco-Hernandez, Daniel Riordan, Joseph Walsh

IMaR Research Centre, Institute of Technology Tralee, Tralee, Ireland
`niall.omahony@research.ittralee.ie`

**Abstract.** Deep Metric Learning (DML) approaches learn to represent inputs to a lower-dimensional latent space such that the distance between representations in this space corresponds with a predefined notion of similarity. This paper investigates how the mapping element of DML may be exploited in situations where the salient features in arbitrary classification problems vary over time or due to changing underlying variables. Examples of such variable features include seasonal and time-of-day variations in outdoor scenes in place recognition tasks for autonomous navigation and age/gender variations in human/animal subjects in classification tasks for medical/ethological studies. Through the use of visualisation tools for observing the distribution of DML representations per each query variable for which prior information is available, the influence of each variable on the classification task may be better understood. Based on these relationships, prior information on these salient background variables may be exploited at the inference stage of the DML approach by using a clustering algorithm to improve classification performance. This research proposes such a methodology establishing the saliency of query background variables and formulating clustering algorithms for better separating latent-space representations at run-time. The paper also discusses online management strategies to preserve the quality and diversity of data and the representation of each class in the gallery of embeddings in the DML approach. We also discuss latent works towards understanding the relevance of underlying/multiple variables with DML.

## 1 Introduction

Deep Learning has great achievements in computer vision for various classification and regression tasks in terms of accuracy, generalisability and robustness. However, to achieve this performance require training on hundreds or thousands of images and very large datasets. Fine-tuning these models for fine-grained visual recognition tasks is not always straightforward however and has prompted the creation of a type of architecture for this type of problem known as metric learning. Metric Learning is popular in

Computer Vision for tasks such as face verification/recognition [1], person re-identification [2, 3], 3D shape retrieval [4] and landmark recognition [5] and is also used in other fields, e.g. for Question Paraphrase Retrieval in Speech Recognition [6], music classification [7] and bioacoustic classification [8] from audio data and gesture recognition from accelerometer data [9]. In Section 2, we will further define the research problems relevant to our research and in Section 3 we will introduce the background theory of Metric Learning for the reader.

The challenges of fine-grained visual recognition relate to two aspects: inter-class similarity and intra-class variance. In Section 4, this paper will review some methodologies which have been proposed in recent research to optimize these two attributes of the embedding space of DML, e.g. through learning dependent relationships in the fields of multi-label classification and newly proposed cost functions, and also methods which exploit the embedding space for interpreting the inner workings of the neural network. In Section 5, this paper will also propose a unique approach to improving classification accuracy of DML in any arbitrary applications through the injection of apriori knowledge of dependent variables into a clustering algorithm appended to the inference pipeline of the DML approach. Examples of such variable features include seasonal and time-of-day variations in outdoor scenes in place recognition tasks for autonomous navigation [5], age/gender variations in human/animal subjects in medical/ethological studies [10] and operator/time-of-shift variations in industrial automation tasks. We will also propose an online management strategy to preserve the quality and diversity of data and the representation of each class in the gallery of embeddings in the DML approach. Finally, in Section 8, this paper will conclude with a discussion of our findings to date and of future work which is currently being actively engaged in follow-up this work.

## 2    Problem Definition

In the field of deep learning, the quality of input data is often more important than the model architecture and training regimen. The challenges of dataset management include ensuring the dataset is correctly labelled, balanced and contains a sufficient amount of data. As well as this, the categories to be classified must also be chosen carefully at the task definition stage to minimize intra-class variance, i.e. it is harder to train a deep learning network to reliably classify 'animals' than it is to train one to classify just 'cats' or 'dogs'. However, breaking down the categories to a low enough level can be difficult, requiring the judgement of an application expert and may introduce unwanted bias. Furthermore, system maintenance does not end once the problem is defined and the model is trained. In situations where salient features to the classification problem vary depending on auxiliary variables, it would be useful to leverage these auxiliary variables (if they are known apriori to classification) to narrow down the classification results to instances which are more likely in light of this new knowledge.

## 2.1 One-Shot Learning

The term One-Shot Learning represents a still-open challenge in computer vision to learn much information about an object category from just one image. Few-shot and zero-shot learning are similar classification problems but with different requirements on how many training examples are available. Few-shot learning, sometimes called low-shot learning often falls under the category of OSL and denotes that multiple images of new object categories are available rather than just one. Zero-shot learning algorithms aim at recognizing object instances belonging to novel categories without any training examples [11]. The motivation for this task lies not only in the fact that humans, even children, can usually generalize after just one example of a given object but also because models excelling at this task would have many useful applications. Example applications include facial recognition in smart devices, person re-identification in security applications as well as miscellaneous applications across industry, e.g. fine-grained grocery product recognition by [13], drug discovery in the pharmaceutical industry [12], stable laser vision seam-tracking systems [13] and the detection of railway track switches, face recognition for monitoring operator shift in railways and anomaly detection for railway track monitoring [14, 15]
.

If it is desired for a conventional machine learning classifier to identify new classes on top of those it was trained to classify then the data for these classes must be added to the dataset (without unbalancing the dataset) and the model must be retrained entirely. This is why metric learning is so useful in these situations where information must be learnt about new object categories from one, or only a few, training samples. The general belief is that gradient-based optimization in high capacity classifiers requires many iterative steps over many examples to perform well. This type of optimization performs poorly in the few-shot learning task.

In this setting, rather than there being one large dataset, there is a set of datasets, each with few annotated examples per class. Firstly, they would help alleviate data collection as thousands of labelled examples are not required to attain reasonable performance. Furthermore, in many fields, data exhibits the characteristic of having many different classes but few examples per class. Models that can generalize from a few examples would be able to capture this type of data effectively.

Gradient descent-based methods weren't designed specifically to perform well under the constraint of a set number of updates nor guarantee speed of convergence, beyond that they will eventually converge to a good solution after what could be many millions of iterations. Secondly, for each separate dataset considered, the network would have to start from a random initialization of its parameters.

Transfer learning can be applied to alleviate this problem by fine-tuning a pre-trained network from another task which has more labelled data; however, it has been observed that the benefit of a pre-trained network greatly decreases as the task the network was trained on diverges from the target task. What is needed is a systematic way to learn a beneficial common initialization that would serve as a good point to start training for the set of datasets being considered. This would provide the same benefits as transfer learning, but with the guarantee that the initialization is an optimal starting point for fine-tuning. [16]

Over years many algorithms have been developed in order to tackle the problem of One-shot learning including:
- Probabilistic models based on Bayesian learning [17, 18],
- Generative models using probability density functions [19, 20],
- Applying transformation to images [21, 22],
- Using memory augmented neural networks [23],
- Meta-learning [16, 24] and
- Metric learning

This paper will focus on the metric learning approach because of the way it learns to map it's output to a latent space and how this may be exploited to infer relationships between feature variability and auxiliary background information.

## 2.2 Fine-Grained Visual Categorization

Fine-grained visual categorization (FGVC) aims to classify images of subordinate object categories that belong to a same entry-level category, e.g., different species of vegetation [25], different breeds of animals [26] or different makes of man-made objects [27].

The visual distinction between different subordinate categories is often subtle and regional, and such nuance is further obscured by variations caused by arbitrary poses, viewpoint change, and/or occlusion. Annotating such samples also requires professional expertise, making dataset creation in real-world applications of FGVC expensive and time-consuming. FGVC thus bears problem characteristics of few-shot learning.

Most existing FGVC methods spend efforts on mining global and/or regional discriminative information from training data themselves. For example, state-of-the-art methods learn to identify discriminative parts from images of fine-grained categories through the use of methods for interpreting the layers of Convolutional Neural Networks, e.g. Grad-CAM [28]. However, the power of these methods is limited when only few training samples are available for each category. To break this limit, possible solutions include identifying auxiliary data that are more useful for (e.g., more related to the FGVC task of interest, and also better leveraging these auxiliary data. These solutions fall in the realm of domain adaptation or transfer learning and the latter has been implemented by training a model to encode (generic) semantic knowledge from the auxiliary data,e.g. unrelated categories of ImageNet, and the combined strategy of pretraining followed by fine-tuning alleviates the issue of overfitting. However, the objective of pre-training does not take the target FGVC task of interest into account, and consequently, such obtained models are suboptimal for transfer. An important issue to achieve good transfer learning is that data in source and target tasks should share similar feature distributions. If this is not the case, transfer learning methods usually learn feature mappings to alleviate this issue.

Alternative approaches include some of those listed for one-shot learning above. Meta-learning has been adopted by [29] to directly identify source data/tasks that are more related to the target one, i.e. select more useful samples from the auxiliary data and remove noisy, semantically irrelevant images. Metric learning has been used similarly during training dataset creation through partitioning training images within each category into a few groups to form the triplet samples across different categories as well as different groups, which is called Group Sensitive TRiplet Sampling (GS-TRS). Accordingly, the triplet loss function is strengthened by incorporating intra-class variance with GS-TRS, which may contribute to the optimization objective of triplet network [27].

Metric Learning has also been employed to overcome high correlation between subordinate classes by learning to represent objects so that data points from the same class will be pulled together while those from different classes should be pushed apart from each other. Secondly, the method overcomes large intra-class variation (e.g., due to variations in object pose) by allowing the flexibility that only a portion of the neighbours (not all data points) from the same class need to be pulled together. The method avoids difficulty in dealing with high dimensional feature vectors (which require $O(d^2)$ for storage and $O(d^3)$ for optimization) by proposing a multi-stage metric learning framework that divides the large-scale high dimensional learning problem to a series of simple subproblems, (achieving $O(d)$ computational complexity) [27].

## 3    Metric Learning

Generally speaking, Metric learning can be summarised by the learning of a similarity function which is trained to output a representation of its input, often called an embedding. During training, an architecture consisting of several identical entities of the network being trained is used along with a loss function to minimize the distance between embeddings of the same class (intra-class variability) and maximize the space between classes (inter-class similarity) so that an accurate prediction can be made. The resulting embedding of each query input is compared using some distance metric against a gallery of embeddings which have been collected from previous queries. In this way, queries need not necessarily be in the training data in order to be re-identified, making the methodology applicable to problems such as facial authentication and person re-identification in security and other one-shot or few-shot learning applications.

Features extracted from classification networks show excellent performance in image classification, detection and retrieval, especially when fine-tuned for target domains. To obtain features of greater usefulness, end-to-end distance metric learning (DML) has been applied to train the feature extractor directly. DML skips the final SoftMax classification layer normally present at the end of CNN's and projects the raw feature vectors to learned feature space and then classifies input image based on how far they are from learned category instances as measured by a certain distance metric.

Due to the simplicity and efficiency, the metric-based approach has been applied in industry for tasks like face recognition and person re-identification [30].

The metric-based methods can achieve state-of-the-art performance in one-shot classification tasks, but the accuracy can be easily influenced when the test data comes from a different distribution [13] The way metric learning works in practice is to have a general model which is good at learning how to represent object categories as 'embeddings', i.e. feature maps, in a feature space such that they all categories are spaced far enough away from each other that they are distinguishable. The second step is to compare each embedding that this model generates for the input image with the embeddings of all previously seen objects. If the two embeddings are close enough in the feature space (shown in Fig. 1) beyond a certain threshold, then the object is identified. The library of embeddings that are compared from may be updated continuously by adding successfully identified embeddings by some inclusion prioritization. If an object is not identified, an external system, e.g. a human expert, may need to be consulted for the correct object label to be applied.

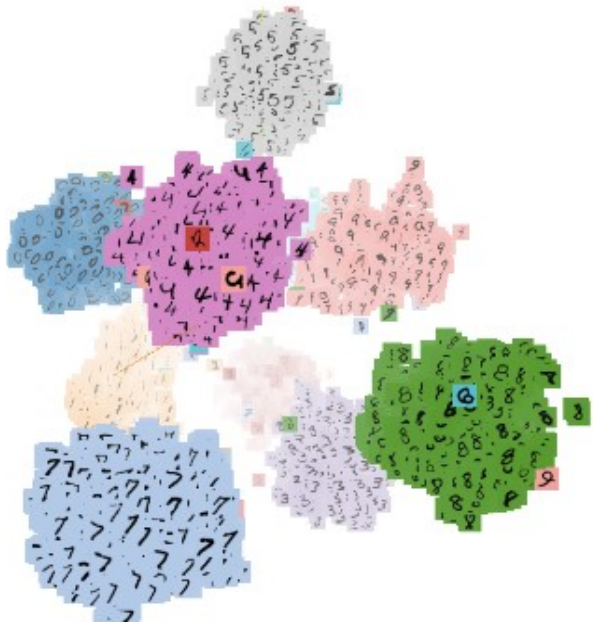

**Fig. 1. A t-SNE (T-distributed stochastic neighbour embedding) visualization of a feature space used in metric learning of the MNIST dataset** [31].

### 3.1 Distance Metrics

Two images, $x_1$ and $x_2$, are compared by computing the distance d between their embeddings $f(x_1)$ and $f(x_2)$. If it is less than a threshold (a hyperparameter), it means that the two pictures are the same object category, if not, they are two different object categories.

$$d(x_1, x_2) = \|f(x_1) - f(x_2)\|$$
(1)

Where $f$ is defined as a parametric function denoting the neural network described earlier that maps high-resolution inputs (images $x_1$ and $x_2$) to low-resolution outputs (embeddings $f(x_1)$ and $f(x_2)$).

It is important to note the distance metric used as this will be used in the loss function which has to be differentiable with respect to the model's weights to ensure that negative side effects will not take place. Distance function which are often used include the Euclidean distance or the squared Euclidean distance [32], the Manhattan distance (also known as Manhattan length, rectilinear distance, L1 distance or L1 norm, city block distance, Minkowski's L1 distance, taxi-cab metric, or city block distance), dot product similarity, Mahalanobis, Minkowski, Chebychev, Cosine, Correlation, Hamming, Jaccard, Standardized Euclidean and Spearman distances [33]

### 3.2 Loss Functions

Loss in metric learning is defined as a measure of the distance of embeddings from sets of similar and dissimilar embeddings. For example, if two images are of the same class, the loss is low if the distance between their associated feature vectors are low, and high if the distance between their associated feature vectors is high. Vice versa, if the two images are of different classes, the loss is only low when the image feature representations are far apart. There are many types of loss function as will become apparent in the next section which will discuss the different kinds of metric learning architecture.

### 3.3 Architectures

There are a number of different ways in which the base feature extractor is embedded in a metric learning architecture. By and large, the general attributes of these architectures include:
a) an ability to learn generic image features suitable for making predictions about unknown class distributions even when very few examples from these new distributions are available
b) amenability to training by standard optimization techniques in accordance with the loss function that determines similarity
c) being unreliant on domain-specific knowledge to be effective.
d) An ability to handle both sparse data and novel data.

To develop a metric learning approach for image classification, the first step is to learn to discriminate between the class-identity of image pairs, i.e. to get an estimate of the probability that they belong to the same class or different classes. This model can then be used to evaluate new images, exactly one per novel class, in a pairwise manner against the test image. The pairing with the highest score according to the network is then awarded the highest probability. If this probability is above a certain threshold then the features learned by the model are sufficient to confirm or deny the identity test image from the set of stored class identities and ought to be sufficient for similar objects, provided that the model has been exposed to a good variety of scenarios to encourage variance amongst the learned features [34].

**Siamese Network**

A Siamese neural network has the objective to find how similar two comparable things are and are so-called as they consist of two identical subnetworks (usually either CNNs or autoencoders), which both have the same parameters and weights. The basic approach of Siamese networks can be replicated for almost any modality.

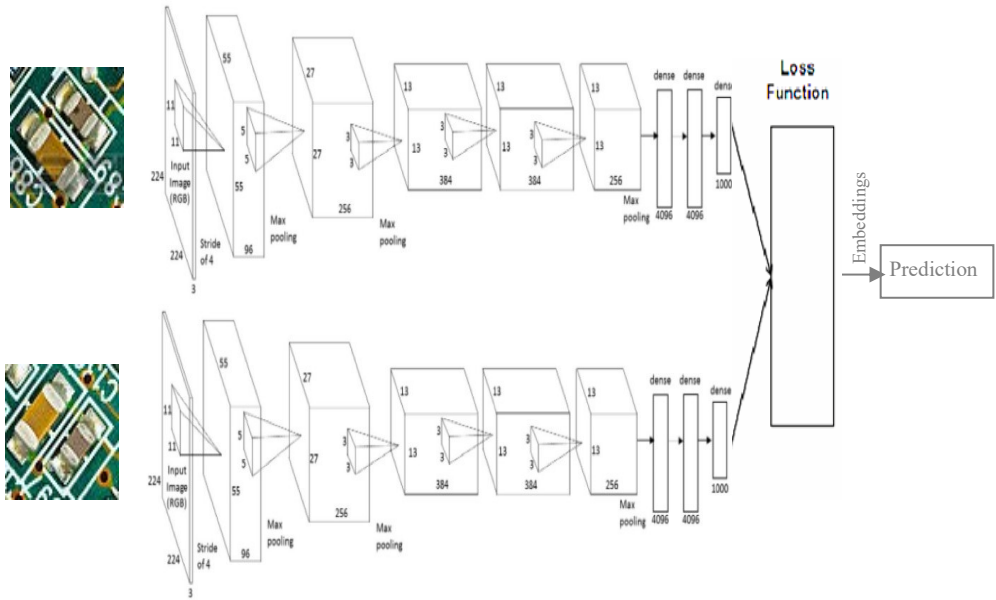

Fig. 2. Siamese Network Architecture

The output of many Siamese networks are fed to a contrastive loss function, which calculates the similarity between the pairs of images ($x_i$ and $x_j$). The input image $x_i$ with samples from both similar and dissimilar sets. For every pair ($x_i$ and $x_j$), if they belong to the set of similar samples S, a label of 0 is assigned to the pair, otherwise, it a label of 1 is assigned. In the learning process, the system needs to be optimized such that the distance function $d$ is minimized for similar images and increased for dissimilar images according to the following loss function:

$$L(x_i, x_j, y) = y \cdot d(x_1, x_2)^2 + (1-y)\max(m - d(x_1, x_2))^2 \qquad (2)$$

**Triplet Network**

The triplet loss is the key to utilize the underlying connections among instances to achieve improved performance. In a similar manner to Siamese networks, triplet networks consist of three identical base feature extractors. The triplet loss function is a more advanced loss function using triplets of images: an anchor image $x_a$, a positive image $x_+$ and a negative image $x_-$, where ($x_+$ and $x_a$) have the same class labels and ($x_-$ and $x_a$) have different class labels. Intuitively, triplet loss encourages to find an embedding space where the distances between samples from the same classes ( i.e., $x_+$ and $x_a$) are smaller than those from different classes ( i.e., $x_-$ and $x_a$) by at least a margin m (Fig. 3). Specifically, the triplet loss could be computed as follows:

$$Ltpl = \sum_{i=1}^{n} \max(0, m + d(x_+, x_a) - d(x_-, x_a)) \quad (3)$$

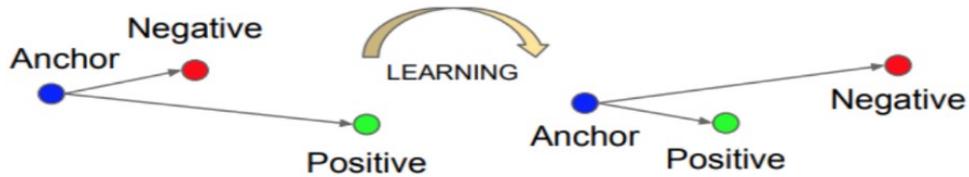

Fig. 3 The Triplet Loss minimizes the distance between an anchor and a positive, both of which have the same identity, and maximizes the distance between the anchor and a negative of a different identity [35].

One advantage of the triplet loss is that it tries to be less "greedy" than the contrastive loss (which considers pairwise examples). The contrastive loss, on the other hand, only considers pairwise examples at a time, so in a sense, it is more greedy. The triplet loss is still too greedy however since it heavily depends on the selection of the anchor, negative, and positive examples. The magnet loss introduced by [36] tries to mitigate this issue by considering the distribution of positive and negative examples. [37] compares these different loss functions and found that End-to-end DML approaches such as Magnet Loss show state-of-the-art performance in several image recognition tasks although they yet to reach the performance of simple supervised learning.

Another popular distance-based loss function is the center loss, which calculated on pointwise on 3d point cloud data. The emerging domain of geometric deep learning is an intriguing one as begin to leverage the information within 3D data. Center loss and triplet loss have been combined in the domain of 3d object detection to be able to achieve significant improvements compared with the state-of-the-art. After that, many variants of triplet loss have been proposed. For example, PDDM [38] and Histogram Loss [39] use quadruplets.

**Quadruplet Network**

The quadruplet network was designed on the intuition that more instances/replications of the base network as shown in Fig. 4) lead to better performance in the learning process. Therefore a new network structure was introduced by adding as many instances into a tuple as possible (including a triplet and multiple pairs) and connect them with a novel loss combining a pair-loss (which connects outputs of exemplar branch and instances branch) and a triplet based contractive-loss (which connects positive, negative and exemplar branches) [29, 40]. Beyond quadruplets, more recent works have used networks with even more instances, such as the n-pair loss [41] and Lifted Structure [39] which place constraints on all images in batches.

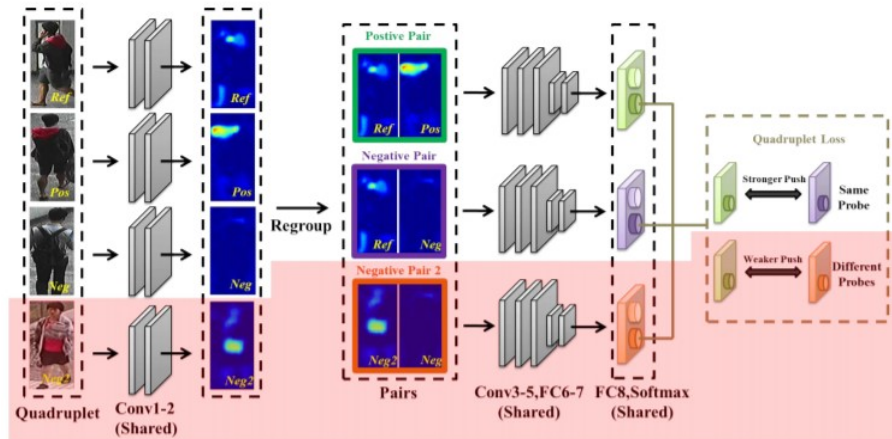

**Fig. 4** Quadruplet Network

### 3.4 The head of the network architecture

The attributes of the network head where the replica base networks meet are also influential on performance. Networks which have been used at this stage include (e.g. which may be a fully-connected layer, a SoftMax layer or a direct throughput.

Another attribute that is controlled at the network head is the level of data augmentation. Data augmentation is a key step to ensuring the model has been exposed to sufficient variance at the training phase that is representative of the real world conditions. By rotating, blurring, or cropping image data, synthetic images can be created that approximately mirror the distribution of images in the original dataset. This method is not perfect, however—it provides a regularizing effect that may be unwanted if the network is already not performing well in training. It is worth noting that training takes significantly longer when data augmentation is applied, e.g. it takes 10 times longer if we apply flip augmentation with 5 crops of each image, because a total of 10 augmentations per image needs to be processed (2 flips times 5 crops).

Another set of hyperparameters is how the embeddings of the various augmentations should be combined. When training using the Euclidean metric in the loss, simply taking the mean is what makes the most sense. But if one, for example, trains a normalized embedding, The embeddings must be re-normalized after averaging at the aggregation stage in the head network. Fig. 5 shows how the network head links these attributes.

**Fig. 5** The metric learning graph in Tensorboard

# 4 Related Work

## 4.1 Dependent Variables

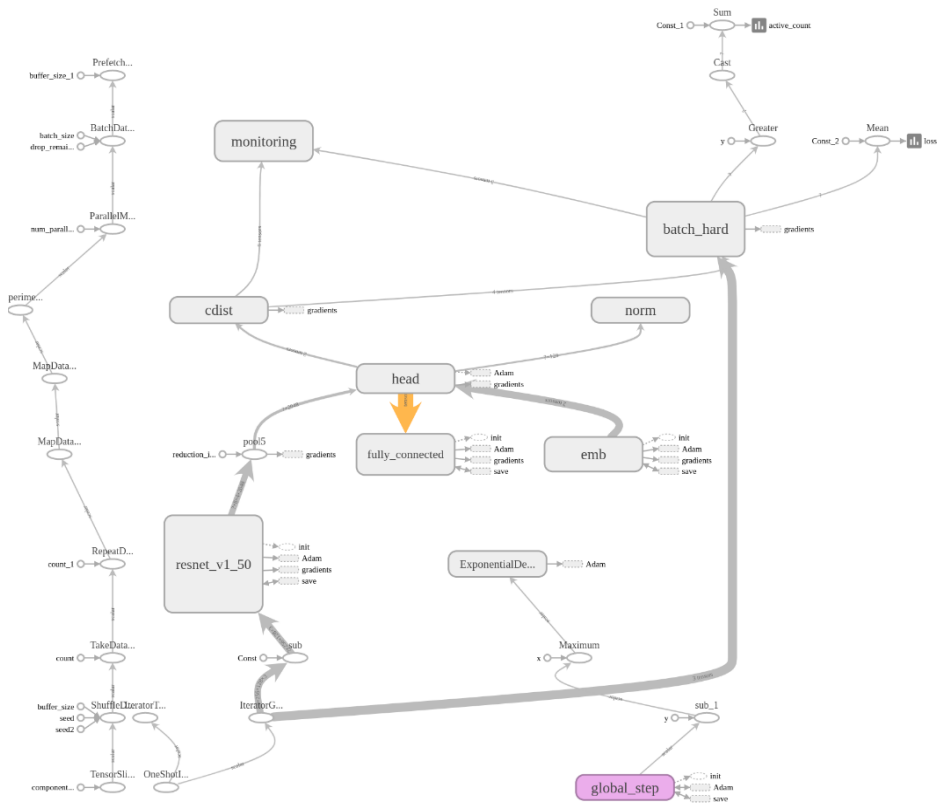

The loss function design in metric learning could be a subtle way of dealing with high degrees of variance due to dependent variables. The contrastive loss pulls all positives close, while all negatives are separated by a fixed distance. However, it could be severely restrictive to enforce such a fixed distance for all negatives. This motivated the triplet loss, which only requires negatives to be farther away than any positives on a per-example basis, i.e., a less restrictive relative distance constraint. However, all the aforementioned loss functions formulate relevance as a binary variable. The use of a ladder loss has been proposed by (Zhou *et al.*, no date) to extend the triplet loss inequality to a more general inequality chain, which implements variable push-away margins according to respective relevance degrees measured by a proper Coherent Score metric.

### 4.2 Multi-label/ Multi-Feature/Muti-Task Learning

Multi-task learning can be seen as a form of inductive transfer which can help improve a model by introducing inductive bias. The inductive bias in the case of multi-task learning is produced by the sheer existence of multiple tasks, which causes the model to prefer the hypothesis that can solve more than one task. Multi-task learning usually leads to better generalization [43]. Multi-label metric learning extends metric learning to deal with multiple variables with the same network. Instances with the more different labels are spread apart, but ones with identical labels will concentrate together. Therefore, introducing more variables means that the latent space is distributed in a more meaningful way in relation to the application domain

It has been proposed in recent work that multiple features should be used for retrieval tasks to overcome the limitation of a single feature and further improve the performance. As most conventional distance metric learning methods fail to integrate the complementary information from multiple features to construct the distance metric, a novel multi-feature distance metric learning method for non-rigid 3D shape retrieval which can make full use of the complementary geometric information from multiple shape features has been presented [4].

An alternative formulation for multi-task learning has been proposed by [44] who use a recent version of the K Nearest Neighbour (KNN) algorithms(large margin nearest neighbour) but instead of relying on separating hyperplanes, its decision function is based on the nearest neighbour rule which inherently extends to many classes and becomes a natural fit for multi-task learning [44]. This approach is advantageous as the feature space generated from Metric Learning crucially determines the performance of the KNN algorithm, i.e. the learned latent space is preserved, KNN just solves the multi-label problem within.

## 5 Our Approach

### 5.1 Using the Latent Space to understand Dependent Variables

Often the feature vector or embedding output is a 128 x1 vector or something of that order meaning that the latent space has 128 dimensions and therefore impossible for humans to visualise. There are tools, however, for dimensionality reduction of the latent space, e.g. PCA (Principal Component Analysis) and t-SNE (T-distributed stochastic neighbour embedding) are available on Tensorboard [45]

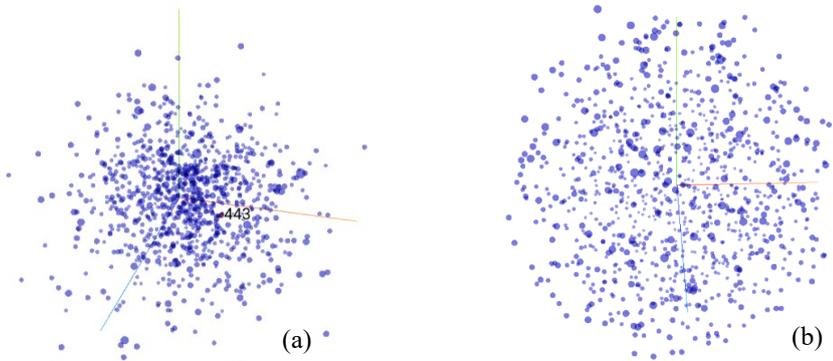

**Fig. 6** (a) PCA (Principle Component Analysis) and (b) t-SNE (T-Distributed Stochastic Neighbour Embedding) projections to 3 dimensions of a latent space with 1024 embeddings.

Many works have used these visualisation tools to interpret the performance of the DML model [29], as well as breakdown attributes of the input relevant to the application as demonstrated by [46] who map transient scene attributes a small number of intuitive dimensions to allow characteristics such as level of snow/sunlight/cloud cover to be identified in each image of a scene.

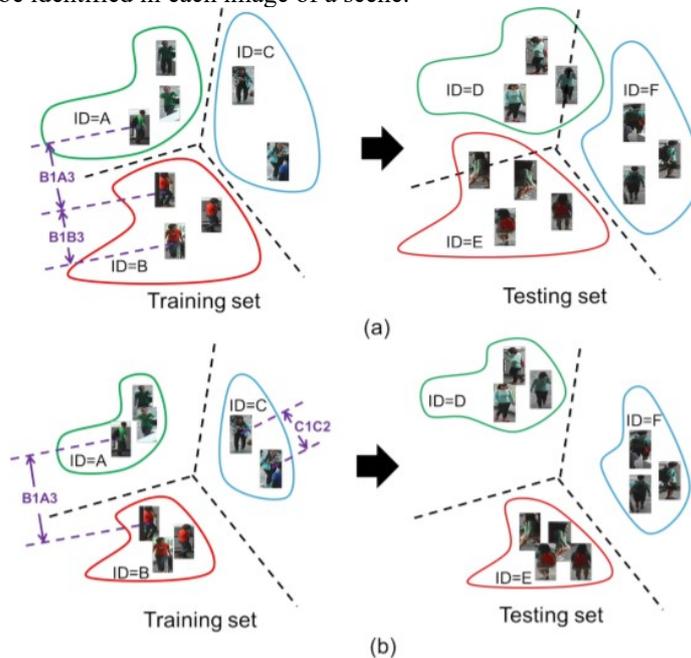

**Fig. 7** A Visualisation with images corresponding to each embedding as shown here in work comparing the performance of (a) triplet loss and (b) quadruplet loss and assess attributes such as l intra-class variation and a large inter-class variation [29].

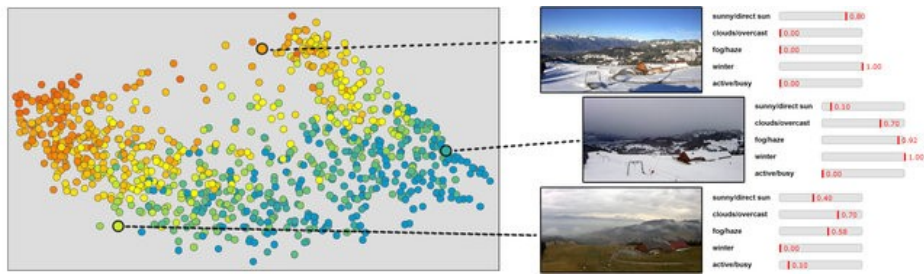

**Fig. 8** Embeddings may also be colourised according to the state of background variables, revealing distributions in the latent space which can lead to better understandings and inference results.

### 5.2 Clustering in the latent space based on Auxiliary background variables

In situations where salient features to the classification problem vary depending on auxiliary variables, it would be useful to leverage these auxiliary variables (if they are known apriori to classification) to narrow down the classification results to instances which are more likely in light of this new knowledge. Better still, if a clustering algorithm, e.g. k-means clustering, could be formulated taking as input the salient background variables and outputting a function which maps the latent space to valid classifications. For specificity, we take the example of the cross-season correspondence dataset [46]. As depicted in Fig. 9, this dataset could be used in future work to prove our proposition that clustering the latent space according to the known time of year may be used to minimise the inter-class similarity to below the acceptable threshold, $\tau$, used at the classification stage.

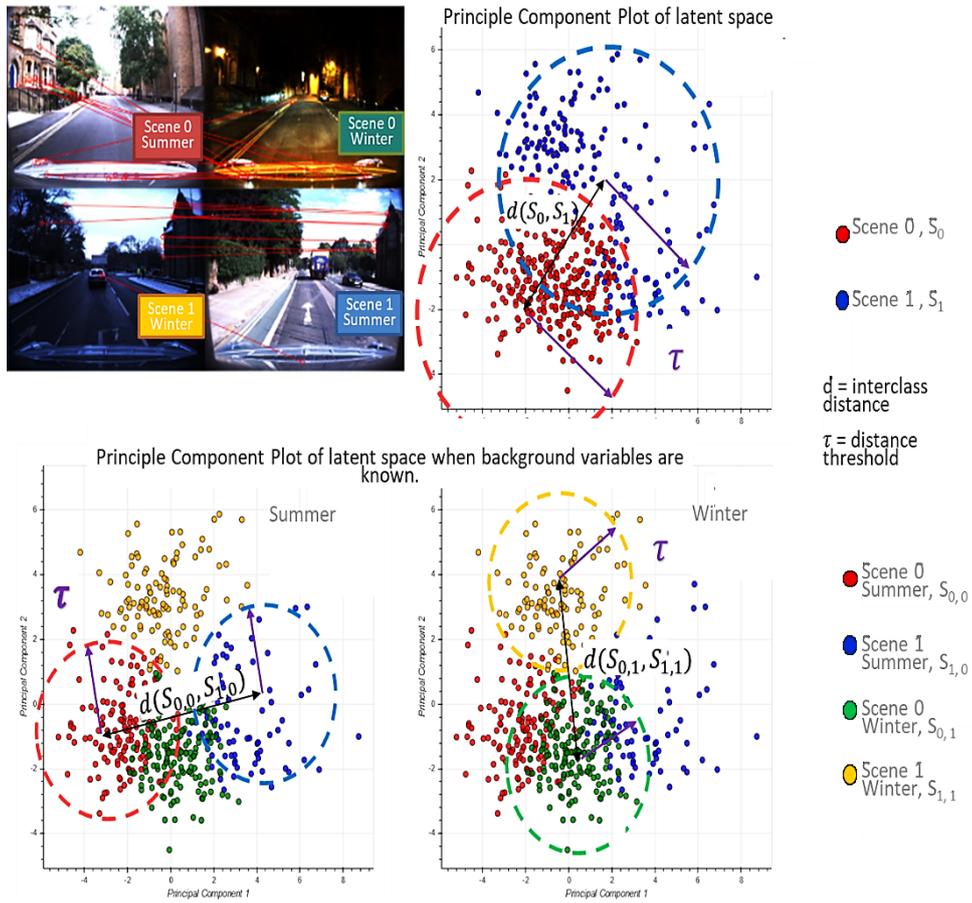

**Fig. 9** A PCA projection of the latent space in DML showing how priori knowledge of background variables, e.g. seasonal variations in outdoor scenes in place recognition, may be used to minimize the intra-class variance and inter-class similarity such that the distance threshold, $\tau$, is less than the distance between classes, $d(S\_0,0, S\_1,0)$.

### 5.3 Gallery Management

We propose that a function to select all embeddings for each class, delete old embeddings given there are more than N (an arbitrary number which may change based on performance results) embeddings for a class and then to compute and remove outliers by some method, e.g. Median Absolute Deviation (MAD) that the representativity of the gallery embeddings of the ground truth, and hence classification accuracy could be improved.

The embeddings are typically written into the HDF5 file in many of the GitHub repositories of previous work. This file format is useful for accessing large amounts of data quickly, however, it does not facilitate the removal of data entries as is desired, e.g. for removing old/noisy embeddings from the gallery over time.

Also, the integration of adaptive thresholding [47] or deep variational metric learning [48] which are methods which allow the distance threshold under which query embeddings must be from embeddings in the gallery to be classified variant to the distribution of embeddings could improve results even more substantially with our proposed method for gallery maintenance.

# 6 Acknowledgement

This work was supported, in part, by Science Foundation Ireland grant 13/RC/2094 and co-funded under the European Regional Development Fund through the Southern & Eastern Regional Operational Programme to Lero - the Irish Software Research Centre ( www.lero.ie )

# 7 Conclusion

This paper investigates how the mapping element of DML may be exploited in situations where the salient features in arbitrary classification problems vary dependent on auxiliary background variables. Through the use of visualisation tools for observing the distribution of DML representations per each query variable for which prior information is available, the influence of each variable on the classification task may be better understood. Based on these relationships, prior information on these salient background variables may be exploited at the inference stage of the DML approach by using a clustering algorithm to improve classification performance. This research proposes such a methodology establishing the saliency of query background variables and formulating clustering algorithms for better separating latent-space representations at run-time. The paper also discusses online management strategies to preserve the quality and diversity of data and the representation of each class in the gallery of embeddings in the DML approach. We also discuss latent works towards understanding the relevance of underlying/multiple variables with DML.

## 7.1 Future Work

Performance comparison with existing not been achieved in this investigation work, however, the concept has promising future results, and the obvious next step in this investigation is to implement our approach on a publically available dataset to ensure reproducibility. The implementation of the proposed solution may be performed, for example, using the 3DWF dataset which contains demographic data such as age or gender is provided for every subject of a face dataset. By taking age, gender and ethnicity as the desired output variables in a multi-task metric learning approach primarily aimed at age estimation from 3D face data. We propose to project the discovered latent space to a representation with dimensions/directions for age, gender and ethnicity. In this way, we may demonstrate how our approach may be used to interpret relationships between binary, ordinal, continuous and seemingly nominal variables.

User interface could be the difference between powerful machine learning tools being a black box that may or not be trusted or a cognitive tool that extends human capabilities at understanding complicated data streams. Reasoning about data through representations can be useful even for kinds of data we understand well because it can make explicit and quantifiable things that are normally tacit and subjective. We propose that the latent space occupied by the representation discovered by metric learning may be exploited.